\documentclass[letterpaper]{article} 
\usepackage{aaai25}  
\usepackage{times}  
\usepackage{helvet}  
\usepackage{courier}  
\usepackage[hyphens]{url}  
\usepackage{graphicx} 
\usepackage{natbib}  
\usepackage{caption} 
\usepackage{algorithm}
\usepackage{algorithmic}

\usepackage{enumitem}

\usepackage{amsmath}

\usepackage{pifont}
\newcommand{\cmark}{\text{\ding{51}}}
\newcommand{\xmark}{\text{\ding{55}}}

\usepackage{listings}
\usepackage{tcolorbox}

\newcommand{\p}{{$p$}}

\newtcolorbox{examplebox}[2][]{fontupper=\tiny,fontlower=\tiny,colback=gray!5!white, colframe=gray!75!black,
    fonttitle=\small\bfseries, title=#2,
    #1}

\usepackage{booktabs}
\usepackage{multirow}
\usepackage{xcolor}
\usepackage{colortbl}
\usepackage{amsmath}
\usepackage{amsfonts}
\usepackage{amssymb}

\usepackage{newfloat}
\DeclareCaptionStyle{ruled}{labelfont=normalfont,labelsep=colon,strut=off} 
\floatstyle{ruled}
\newfloat{listing}{tb}{lst}{}
\floatname{listing}{Listing}
\title{Trustworthy and Practical AI for Healthcare: A Guided Deferral System with Large Language Models}
\author{
    Joshua Strong,
    Qianhui Men\footnote{Work undertaken whilst employed at the University of Oxford.}, 
    J. Alison Noble
}
\affiliations{
    Institute of Biomedical Engineering, Department of Engineering Science, University of Oxford, OX3 7DQ, United Kingdom\\
    \{joshua.strong, alison.noble\}@eng.ox.ac.uk, qianhui.men@bristol.ac.uk
}

\begin{document}

\maketitle

\begin{figure*}[t]
  \centering
  \includegraphics[width=0.8\textwidth]{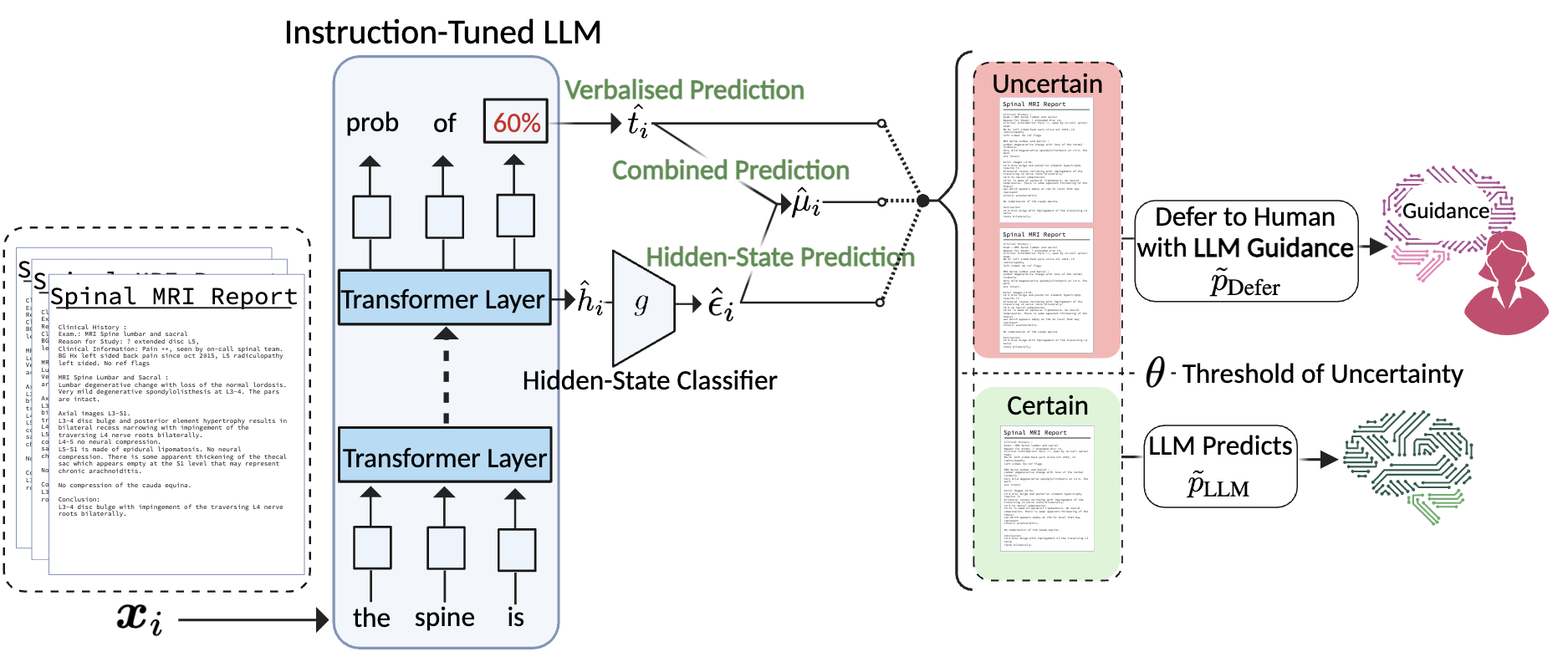}
  \caption{Our \textit{guided deferral} system. Reports are parsed by an instruction-tuned LLM for clinical disorders. From the text output, we extract a \textit{verbalised prediction} $\hat{t}$. We calculate a \textit{hidden-state} $\hat{\epsilon}$ prediction from the final hidden-layer of the LLM, and its combination with $\hat{t}$ through their mean $\hat{\mu}$. Uncertain predictions, determined by either $\hat{t}$, $\hat{\epsilon}$, or $\hat{\mu}$, are deferred to humans with guidance. Certain predictions are autonomously handled by the LLM. Created in BioRender. Strong, J. (2025).}
  \label{paper_diag}
\end{figure*}

\begin{abstract}
Large language models (LLMs) offer a valuable technology for various applications in healthcare. However, their tendency to hallucinate and the existing reliance on proprietary systems pose challenges in environments concerning critical decision-making and strict data privacy regulations, such as healthcare, where the trust in such systems is paramount. Through combining the strengths and discounting the weaknesses of humans and AI, the field of Human-AI Collaboration (HAIC) presents one front for tackling these challenges and hence improving trust. This paper presents a novel HAIC guided deferral system that can simultaneously parse medical reports for disorder classification, and defer uncertain predictions with intelligent guidance to humans. We develop methodology which builds efficient, effective and open-source LLMs for this purpose, for the real-world deployment in healthcare. We conduct a pilot study which showcases the effectiveness of our proposed system in practice. Additionally, we highlight drawbacks of standard calibration metrics in imbalanced data scenarios commonly found in healthcare, and suggest a simple yet effective solution: the Imbalanced Expected Calibration Error.
\end{abstract}

%
\begin{links}
    \link{Code}{https://github.com/josh-strong/Guided_Defer_LLM}
\end{links}

\section{Introduction}

\label{Introduction}
The practical implementation of artificial intelligence (AI) in decision-sensitive environments such as healthcare is often a delicate trade-off between the benefits of autonomy, risk, and the costly consequences of errors. Human-AI Collaboration (HAIC), whereby humans and AI work in tandem to produce outcomes superior to independent efforts, seeks to find this balance \cite{wang2020human}. One method of achieving HAIC is through utilising \textit{deferral systems} \cite{madras2018predict,liu2022incorporating}, which aim to identify cases that can be rapidly and accurately processed by the AI and then defer the remaining challenging cases to humans. This collaboration reduces decision-makers' time and fatigue, enabling them to focus on a small set of uncertain cases identified by the deferral system, with confidence that the AI will efficiently and accurately process the majority of cases. By ensuring that AI handles routine tasks while humans manage the complex ones, the system not only optimises the balance between benefits and risks, but crucially can also enhance human trust in the AI’s role.

Clinicians often make decisions using their expertise in addition to the \textit{intelligent guidance} from colleagues. The colleagues will then respond with their interpretation and the reasoning behind it. However, existing deferral systems lack this guidance \cite{banerjee2023learning}, isolating humans in this critical decision-making. In the spirit of HAIC, we posit that effective deferral systems should additionally offer intelligent guidance to human decision-makers in deferred cases. In this paper, we novelly explore utilising large language models (LLMs) for this purpose, specifically in the task of clinical text parsing for disorder classification.

The main challenges in deploying such systems in practice include the computational cost of LLMs and compliance with strict data privacy regulations \cite{freyer2024future}. Proprietary LLMs offer advantages such as state-of-the-art performance and easy implementation without requiring high-performance hardware, but are impractical for data-sensitive applications such as healthcare due to lack of internal state access and privacy concerns \cite{ullah2024challenges}. Open-source LLMs can perform well with relatively large amounts of parameters, but are slow and require high-performance hardware, not readily available to all practitioners. Small-scale LLMs are faster and can be run on widely available hardware, but are not effective. 
A closely-related study \cite{zöller2024humanaicollectivesproduceaccurate} that successfully applied LLM for medical record diagnosis has ensembled multiple LLMs and human physicians. Rather than deferring cases to humans, it focuses on a hybrid collaborative approach which may cause cognitive overload.
In this paper, we propose methodology for developing \textit{efficient}, \textit{effective} and \textit{open-source} LLMs capable of guided deferral, suitable for healthcare applications and deployable on modest computational resources.

The performance of deferral systems can be attributed, in part, to the model's classification performance and its \textit{calibration}—the ability for the model to accurately reflect its confidence in its predictions. 
Appropriate calibration improves the model's ability to choose the optimal hierarchy of cases to be deferred. Recent literature has highlighted the limitations \cite{nixon2020measuring} of the \textit{Expected Calibration Error} (ECE) \cite{guo2017calibration}, a commonly used metric for calibration. However, the proposed solutions are often inadequate in settings with class imbalance, which is frequently observed in healthcare data. Through analysis on real-world healthcare examples, we propose a new metric: the \textit{Imbalance Expected Calibration Error} ($\mathrm{ECE_{Imb}}$), which adjusts the calibration error by appropriately re-weighting the original calibration error weights in calculating ECE to account for data imbalance. We find that this metric is more suitable for evaluating the calibration of a model in these situations.

In summary, the contributions of this paper are:

\begin{itemize}[itemsep=1pt, nolistsep]
    \item  We propose a novel deferral system, \textit{guided deferral}, for large language models (LLMs) in computer-aided clinical diagnosis. This system not only defers cases to human decision-makers, but also provides \textit{intelligent guidance}. We detail its practical application and demonstrate its efficacy through a pilot study.
    \item 
    We highlight the challenges of calibrating a deferral system with imbalanced data, common in clinical scenarios between diagnosis and negative cases. A new metric, \textit{$\mathrm{ECE_{Imb}}$}, is proposed as a simple yet effective measurement of calibration in such situations.
    \item We address the social impact challenges associated with deploying such systems in practice, by developing methodology for creating efficient, effective, and open-source LLMs capable of guided deferral.
    It is also deployable on modest computational resources and accessible to the community. 
    \item We position our research within the social impact domain of healthcare, addressing challenges of low-count, imbalanced data and data privacy. We propose a human-AI collaboration system crucial for making key decisions and building trust in sensitive fields.
\end{itemize}

\section{Related Work}
\paragraph{Deferral Systems for Healthcare.}
AI systems can defer appropriately to clinicians, which improves accuracy and efficiency in the human-AI collaborative environment for medical support (\citet{Dvijotham2023}). With a focus on system fairness and safety, foundations for advanced deferral algorithms such as Learning-to-Defer \cite{madras2018predict,liu2022incorporating} have also proposed solutions with uncertainty to assess the confidence level in each prediction.

\paragraph{Human-AI Collaboration with LLMs.} Few studies explore the use of LLMs in HAIC. \citet{wiegreffe2022reframing} first examined LLMs for explaining classification decisions using a human-in-the-loop approach to train a filter assessing explanation quality. \citet{Rastogi_2023} used HAIC to audit error-prone LLMs with other LLMs. The closest work is \citet{zöller2024humanaicollectivesproduceaccurate} who research a hybrid collective intelligence system that combines human expertise with the capabilities of LLMs to improve decision-making in medical diagnostics. Our system differs by using LLMs for guided deferrals. \citet{banerjee2023learning} used LLMs to provide textual guidance for clinicians in decision-making on clinical imaging tasks. 
The system does not defer cases, which still burdens decision-makers with the time and fatigue issues. Our work combines LLMs in deferral systems with valuable guidance for decision-makers on deferred cases. 

\paragraph{Selective Prediction of LLMs.} Existing research on LLMs in selective prediction includes methods to measure uncertainty in model responses after generation \cite{varshney2023postabstention}. \citet{chen2023adaptation} proposed improving selective prediction by incorporating self-assessment. \citet{ren2022out} explored detecting out-of-distribution instances in summarisation and translation tasks. Our paper applies selective prediction to deferral systems in clinical parsing. We show that combining the model's internal state with its generated prediction enhances selective prediction without post-generation methods. We provide a comprehensive evaluation of selective prediction performance in clinical classification using real-world data, applied in deferral systems with in-distribution data.

\paragraph{Instruction-Tuning of LLMs.} Whilst there exists literature demonstrating improved zero-shot performance of instruction-tuned (IT) LLMs on unseen tasks \cite{wei2022finetunedlanguagemodelszeroshot}, our work studies this improvement specifically on an in-domain task through the use of \textit{guardrails}. Additionally, there exists works on the use of IT in the medical domain \cite{liu2023radiology}, but these focus on report summarisation. No works have researched the applications of IT LLMs for the use of deferral systems. \citet{zhang2023instruction} provide a recent survey for additional insight in this area.

\paragraph{On the Calibration of LLMs.}
There is a wealth of literature on the calibration of LLMs \cite{geng2023survey}. \citet{lin2022teaching} illustrated the capacity of LLMs to convey the uncertainty of their model answers through verbalisation, specifically in their textual output. Additionally, \citet{xiong2023llms} assessed the calibration of LLMs in relation to their verbalised probabilities—when prompted to output predictions through the generated text. In this paper, we discuss the pitfalls of LLM calibration under imbalanced classes of samples in healthcare, and propose an unbiased metric to evaluate model calibration.

\paragraph{The Role of Cognitive Science in HAIC.}
In the collaboration between humans and AI to achieve a specific goal, the distinct strengths of each enable \textit{complementarity}, a performance surpassing that of either alone and measuring the HAIC system's success \cite{bansal2021does}. Humans adapt to AI by developing \textit{mental models}—cognitive frameworks aiding understanding and interaction. \citet{Andrews2022} comprehensively reviews mental models, including their definition, measurement, and related concepts. \citet{kulesza2012tell} empirically showed that users with better-developed mental models achieved more satisfactory AI results. Research in cognitive science is crucial for developing effective human-AI collaborative systems.

\section{Methods}

\subsection{Sources of Predictions}
 Utilising LLMs in clinical text parsing for disorder classification presents a unique challenge in determining the classification approach. We focus our study on binary classification utilising the top-performing methods of two distinct sources of classifications; one from the internal-states of the LLM and another from the generated textual output. Additionally, we experiment with a third through combining these sources. Next, we formally define these three sources. See Figure \ref{paper_diag} for a visual supplement.

\paragraph{Verbalised Prediction Source.}
The \textit{verbalised probability} is the probability of the positive class extracted from the generated text of the LLM. We denote this probability for input $\boldsymbol{x}_i$ as $\hat{t_i}$. For example, the verbalised probability for the example guidance generated in Figure \ref{box} is $\hat{t}_i = 0.6$.

\paragraph{Hidden-State Prediction Source.}
The \textit{hidden-state} probability is defined based on the hidden representations of LLM to implicitly detect classifications. Inspired by \citet{ren2022out}, which utilises the final-layer hidden embedding of LLMs for out-of-distribution detection, the output embedding $\hat{h}_i \in \mathcal H$ for input $\boldsymbol{x}_i$ is computed as the average of the decoder's final-layer hidden-state vectors $\boldsymbol{g}_{ik} \in \mathbb{R}^{d}$ over all $K$ output tokens with a hidden dimension of $d=5120$ for the small-scale LLM:
\begin{align*}
    &\hat{h}_i = \frac{1}{K} \sum_{k=1}^{K}  \boldsymbol{g}_{ik}^n.
\end{align*}
Then, an MLP $g: \mathcal{H} \mapsto \mathbb{R}$ is trained as a hidden-state classifier to learn the probability of the disorder from the LLM hidden representation, the \textit{hidden-state prediction probabilities}, denoted $\hat{\epsilon}_i$, i.e.  $\hat{\epsilon}_i = g(\hat{h}_i)$. We experimented with different models for the hidden-state classifier in ambitions of fully utilising the information embedded in $\hat{h}$ (see Technical Appendix for details).

\paragraph{Combined Prediction.}
We combine $\hat{t}_i$ and $\hat{\epsilon}_i$ through a weighted mean to form the \textit{combined prediction}, of which we denote $\hat{\mu}_i$:
\begin{equation*}
    \hat{\mu}_i = \alpha \hat{\epsilon}_i + (1-\alpha) \hat{t_i},
\end{equation*}
where $\alpha \in [0, 1]$ can be learned on a validation data split. For simplicity, we fix $\alpha=0.5$ in our experiments.

\subsection{Instruction-Tuning Methodology}
\begin{figure}[h]
    \centering
    \includegraphics[width=\columnwidth]{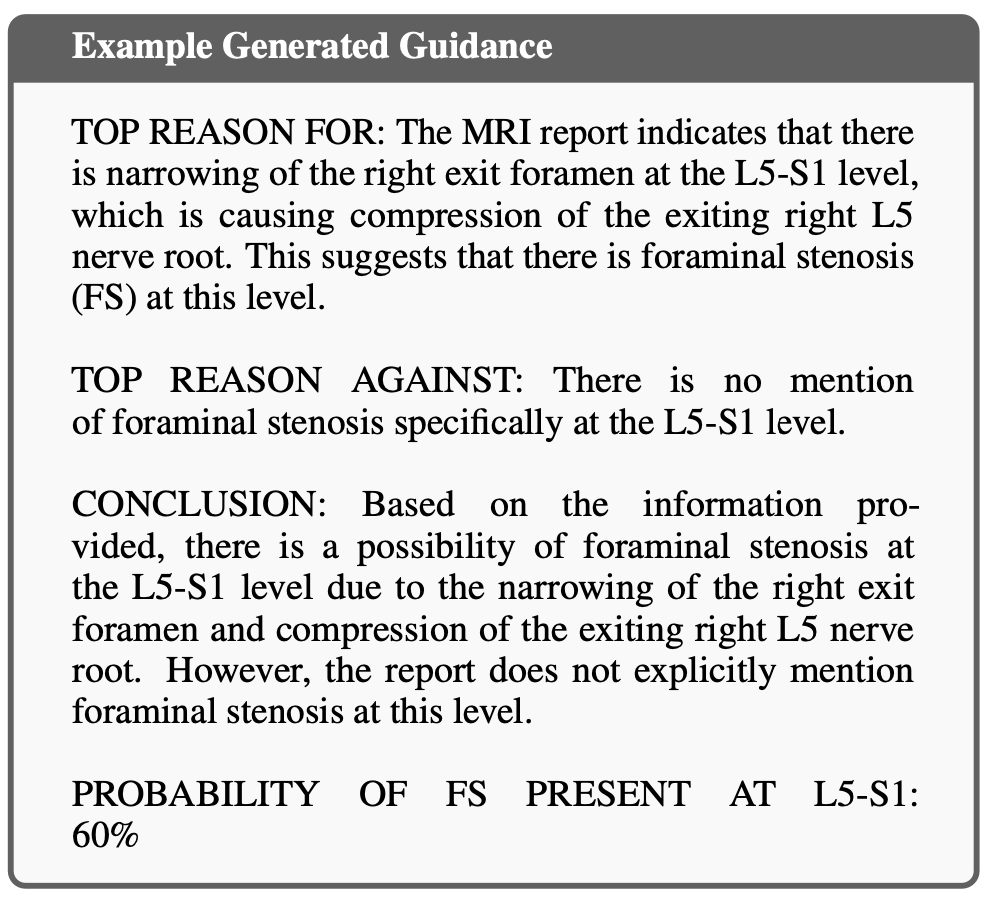}
    \caption{Example guidance based on a spinal MRI report. The instruction-tuned LLM is able to intelligently infer a diagnosis with sound logic without explicit textual diagnosis.}
    \label{box}
\end{figure}

In generating well-calibrated verbalised probabilities, we use the ``\textit{verb. 1S top-$k$}" prompting strategy \cite{tian2023just}, prompting the LLM to provide the top $k$ guesses and their probabilities in a single response. Adapting this to $k=1$ for a binary setting, we prompt the LLM to return the probability of the positive class. Additionally, we prompt for the top reason the disorder \textit{might} and \textit{might not} be present, using \textit{dialectic reasoning} \cite{Hegel_2018} to provide intelligent guidance for decision-makers. This technique has proven effective in decision support \cite{JARUPATHIRUN20071553}. An example output is in Figure \ref{box}.
\paragraph{Instruction-Tuning Data Generation.}
\label{ig} We use 4-bit quantized 13B and 70B versions of the open-source LLM \textit{Tulu V2 DPO} \cite{ivison2023camels}, a Llama 2 \cite{touvron2023} derivative, as our \textit{small} and \textit{large}-scale LLMs respectively.
As classification, calibration and deferral performance was seen to improve with model size (see Table \ref{clf_performance}), we instruction-tune the small-scale models on \textit{guard-railed} answers generated by the large-scale model. We experiment instruction-tuning using multiple generated answers for each example in the instruction-data through enabling the stochastic decoding of the LLM. We \textit{guard-rail} the instruction-tuning data generation process such that answers containing hallucinations and ``bad" behaviours were rejected. Hallucinations are identified when the generated answers did not match the underlying annotation, and ``bad" behaviours when the LLM did not output English language, concise answers, or logical statements. The full implementation of guard-rails and the final base-prompt used to generate these examples can be seen in the Technical Appendix.

\paragraph{Instruction-Tuning.} We instruction-tune the small-scale LLM using QLoRA \cite{dettmers2023qlora} solely on completions. QLoRA enables fine-tuning of LLMs with much less memory requirements whilst maintaining full 16-bit fine-tuning task performance, aligning with our ambitions of a computationally-efficient system. See Technical Appendix on experimental details for full reproducibility.

\subsection{Calibration Methods}
\label{cali_methods}
In improving the deferral performance of our LLM, we must also focus on the calibration of the model in tandem with the classification performance. Note that a highly predictive model can still be poorly calibrated \cite{wang2023calibration}. We motivate and detail our proposed metric \textit{Imbalanced ECE} in this subsection.

\paragraph{Model Calibration.}
 Intuitively, calibration measures a model's ability to accurately quantify its confidence in its predictions through its estimated class probabilities. Let $x$ be a realisation of data $X \in \mathcal{X}$ with corresponding label $y \in Y = \{0,1\}$. A model $h(X): X \rightarrow [0,1]$ producing confidence probabilities $\hat{p}$ of belonging to the positive class is said to be perfectly calibrated  \cite{guo2017calibration} iff
\begin{equation*}
\label{perf_cal}
    P(Y=1 \,|\, h(X)=\hat{p}) = \hat{p}, \quad \forall \hat{p}\in [0,1].
\end{equation*}

\paragraph{Metrics of Calibration.}
A common metric for quantifying the overall calibration of a model is the \textit{Expected Calibration Error} (ECE) \cite{Naeini2015-lh}. Model confidence probabilities $\hat{p}$ are binned into $M$ equal-sized interval bins within the domain $[0,1]$. Let $B_m$ be the set of $\hat{p}$ which fall within the bin bounded by $I_m = (\frac{m-1}{M}, \frac{m}{M}]$, $m \in \{1,...,M\}$. Formally, with a dataset $\{x_i, y_i\}_{i=1}^{n}$, 
\begin{equation*}
    \mathrm{ECE}=\sum_{m=1}^M \frac{\left|B_m\right|}{n}\left|\bar{h}_{m} - \bar{y}_m \right|,
\end{equation*}
where $\bar{h}_{m}$ is the average confidence probability of bin $m$ and $\bar{y}_m$ the average label in bin $m$. The ECE has been shown to have an inherent statistical bias, particularly for extremely well-calibrated models \cite{roelofs2022mitigating}, in addition to sensitivity to hyperparameters, such as the number of bins, type of norm used, and binning technique \cite{nixon2020measuring}. We fix these hyperparameters across experiments in this paper, with the number of bins set to $M=10$ and utilising the $\ell_1$ norm in all calibration metrics.

The \textit{Adaptive Calibration Error} (ACE) was proposed by \citet{nixon2020measuring} as a solution to the problem presented by the ECE, in that the methodology of selecting equally-sized interval bins could result in bins with low-count predictions, and hence, a poor estimate in overall calibration error. Instead, ACE chooses bins $B_{m*}$ in a way such that each of the $M$ bins contains an equal number of $\frac{1}{M}$ confidence probabilities and bin $B_{m*}$ contains the $m*^{th}$ set of sorted predictions:
\begin{equation*}
    \mathrm{ACE}=\sum_{m*=1}^M \frac{1}{M}\left|\bar{h}_{m*} - \bar{y}_{m*} \right|.
\end{equation*}

\paragraph{Imbalanced ECE.}
For imbalanced datasets with a majority negative class, a large proportion of the confidence probabilities of the negative class will inherently lie within the first few bins. In its computation of a weighted average of errors across bins, ECE assigns minimal weights to subsequent bins with higher probability predictions and fewer data points. As a result, the calibration of the majority class overwhelmingly influences the metric, potentially resulting in a misleading interpretation of total calibration through the ECE. The ACE was designed to combat this, but it also suffers in imbalanced situations, as generating equally spaced bins leads to the majority of bins inherently covering only the negative class examples. This problem is further exacerbated by the sharpness of model predictions, a prominent characteristic of neural networks \cite{Gneiting2007-bk}, pushing confidence predictions to their extremities of 0 or 1.

In addressing this issue, \citet{guilbert2024calibration} introduced ECE${@k\%}$, which assesses the ECE of the top $k\%$ most confident predictions by partitioning the probability prediction domain $[\min(\hat{p}),1]$ into $\log{10}(k\%)$ bins. While valuable in scenarios that depend on the high calibration of high-confidence predictions, this metric falls short in evaluating the calibration of all predictions—a critical aspect, especially in deferral system use cases. 

Hence, we introduce a simple yet effective metric \textit{Imbalanced Expected Calibration Error}, $\mathrm{ECE_{Imb}}$, which weights the calibration error by incorporating a blend of both the bin sample proportion and a uniform distribution. Formally, 
\begin{align*}
    \mathrm{ECE_{Imb}} &= \sum_{m=1}^M \Gamma_m \left| \bar{h}_m - \bar{y}_m \right|, \\
    \text{where} \quad \Gamma_m &= (1-\gamma) \frac{|B_m|}{n} + \gamma \frac{1}{M}, \ \gamma \in [0,1].
\end{align*}
When $\gamma=0$, $\mathrm{ECE_{Imb}}$ reduces to the classic $\mathrm{ECE}$. As $\gamma\rightarrow1$, $\Gamma_m$ morphs towards a uniform distribution. $\gamma$ can be heuristically chosen on a validation dataset. In selecting a value for $\gamma$, it is recommended to choose a conservative value which does not draw the weights too much from their original bin-weighted distribution. We further address the issues with ECE, ACE, and the $\mathrm{ECE_{Imb}}$ solution, as well as the logic behind our choice of $\gamma=0.3$ for our experiments, in the Technical Appendix.

\subsection{Deferral Mechanism}
Our deferral mechanism is based on the confidence of predictions, determined by their distance from the decision boundary of 0.5. For a given set of \( n \) predictions \( \hat{p} \in \{ \hat{t}, \hat{\epsilon}, \hat{\mu} \} \), we transform these into sorted relative confidence probabilities \( \tilde{p} \):
\begin{equation*}
\tilde{p} = \text{sort} \left( \{2 \left| \hat{p}_i - 0.5 \right| \}_{i=1}^{n} \right) = \{\tilde{p}_1, \tilde{p}_2, \ldots, \tilde{p}_n\}
\end{equation*}

This transformation allows for an equal comparison between positive and negative classes. The resulting set \( \tilde{p} \) measures uncertainty, where \( \tilde{p}_i \) has higher deferral priority over \( \tilde{p}_j \) if \( \tilde{p}_i < \tilde{p}_j \) for all \( i, j \in \{1, \ldots, n\} \) with \( i \neq j \).

Deferral performance is evaluated by iterating through this hierarchy of \( \tilde{p} \), deferring each prediction without replacement, and measuring the classification performance of the LLM on the remaining cases with the respective \( \hat{p} \) predictions. Note that \( \hat{p} \) and \( \tilde{p} \) do not have to be derived from the same set of predictions \( \hat{t} \), \( \hat{\epsilon} \), or \( \hat{\mu} \); classification and deferral can use separate sources of probabilities. This procedure describes the AUARC (Area Under the Accuracy-Rejection Curve) metric \cite{pmlr-v8-nadeem10a} when measuring accuracy. Good deferral behaviour displays a strictly monotonic increasing accuracy with an increasing deferral rate. In practice, the proportion of cases deferred is subject to a threshold of uncertainty \( \theta \in [0,1] \), dependent on operational constraints and/or risk appetite. Deferred cases $\tilde{p}_{\text{Defer}}$ are those where \( \tilde{p}_i < \theta \), and cases automatically parsed by the LLM $\tilde{p}_{\text{LLM}}$ are those where \( \tilde{p}_i \geq \theta \), formally:
\begin{equation*}
   \tilde{p}_{\text{Defer}} = \{\tilde{p}_i \in \tilde{p} \mid \tilde{p}_i < \theta \} \,\,\, \text{and} \,\,\, \tilde{p}_{\text{LLM}} = \{\tilde{p}_i \in \tilde{p} \mid \tilde{p}_i \geq \theta \}, 
\end{equation*}
such that $\tilde{p}_{\text{Defer}} \cap \tilde{p}_{\text{LLM}} = \emptyset$ and $\tilde{p}_{\text{Defer}} \cup \tilde{p}_{\text{LLM}} = \tilde{p}$.

\subsection{Pilot Study}
To evaluate the effectiveness of our LLM guidance in human decision-making, we conduct a pilot study with 20 participants in the scenario of deferring 30 ($\approx$5\%) of the most uncertain test predictions, such that $\theta = \tilde{p}_{31}$. The participants are researchers with biomedical backgrounds but not specialised in spinal disorders. The 30 medical reports are from OSCLMRIC dataset (introduced in Sect.~\ref{sec:experiments}) with ethical approval obtained.

\paragraph{Study Design.} All participants received background information, including clinical details, examples of prediction outcomes with associated MRI reports and the LLM's performance on a validation set to help participants develop a sufficient mental model of the LLM, of which has been shown to be important in HAIC systems \cite{kulesza2012tell,bansal2019beyond}. The ordering of the 30 questions were randomised for each participant to reduce order bias.
Participants moved to the next question if their prediction matched the LLM's. If it differed, they received guidance and could either change their prediction or keep it based on the guidance and their understanding of the LLM. This process allowed us to assess human performance with and without guidance. Effective guidance should help participants recognise both the accuracy and inaccuracy of their judgements. Full details are in the Technical Appendix.

\section{Experiments}
\label{sec:experiments}

\begin{table*}[!ht]
\centering
\footnotesize
\begin{sc}

\begin{tabular}{lcccccccc}
  \toprule
  \multirow{2}{*}{\textbf{Setup}} & \multicolumn{3}{c}{\textbf{Classification Perf. }} & \multicolumn{3}{c}{\textbf{Calibration Perf.}} & \multicolumn{1}{c}{\textbf{Deferral Perf.}} \\
  \cmidrule(lr){2-4} \cmidrule(lr){5-7} \cmidrule(lr){8-8} &
    \textbf{Recall}$\uparrow$ & \textbf{Precision}$\uparrow$ & \textbf{F1-score}$\uparrow$  & \textbf{ECE}$\downarrow$ & \textbf{ACE}$\downarrow$ & $\textbf{ECE}_{\textbf{Imb}}$$\downarrow$ & \textbf{AUARC}$\uparrow$ \\
   \midrule

  (1) \hspace{0.1cm} $\hat{t}_{\text{BASE-13B}}$ & 0.85 ± 0.02 & 0.61 ± 0.01 & 0.61 ± 0.01 & 0.26 ± 0.01 & 0.32 ± 0.01 &  0.30 ± 0.01  & 0.897460 ± 0.0063 \\

   (2) \hspace{0.1cm} $\hat{\epsilon}_{\text{BASE-13B}}$ & 0.73 ± 0.02 & 0.94 ± 0.03 & 0.80 ± 0.02 & 0.03 ± 0.01 & \textbf{0.04 ± 0.01} & 0.09 ± 0.01 & 0.994715 ± 0.0006 \\

   (3) \hspace{0.1cm} $\hat{\mu}_{\text{BASE-13B}}$ & 0.78 ± 0.03 & 0.87 ± 0.04 & 0.82 ± 0.03 & 0.13 ± 0.01 & 0.18 ± 0.01 & 0.13 ± 0.01 & 0.993318 ± 0.0010 \\
   \midrule

   \addlinespace[4pt] 

   \rowcolor{gray!15} (4) \hspace{0.1cm} $\hat{t}_{\text{BASE-70B}}$ & 0.90 ± 0.02 & 0.78 ± 0.01 & 0.83 ± 0.01 & 0.27 ± 0.01 & 0.33 ± 0.01 & 0.25 ± 0.01 & 0.973529 ± 0.0052 \\

  \rowcolor{gray!15} (5) \hspace{0.1cm} $\hat{\epsilon}_{\text{BASE-70B}}$ & 0.87 ± 0.01 & 0.97 ± 0.01 & 0.91 ± 0.01 & 0.02 ± 0.01 & 0.05 ± 0.01 & \textbf{0.08 ± 0.01} & 0.997191 ± 0.0002 \\

  \rowcolor{gray!15} (6) \hspace{0.1cm} $\hat{\mu}_{\text{BASE-70B}}$ & 0.90 ± 0.01 & 0.95 ± 0.02 & 0.92 ± 0.01 & 0.14 ± 0.01 &0.19 ± 0.01 & 0.14 ± 0.01 & 0.997260 ± 0.0003 \\

   \addlinespace[4pt] 

    \midrule
  (7) \hspace{0.1cm} $\hat{t}_{\text{IT-13B}}$ & 0.85 ± 0.02 & 0.92 ± 0.02 & 0.88 ± 0.02 & 0.27 ± 0.01 & 0.32 ± 0.01 & 0.23 ± 0.01 & 0.987501 ± 0.0024 \\

  (8) \hspace{0.1cm} $\hat{\epsilon}_{\text{IT-13B}}$ & 0.89 ± 0.01 & \textbf{0.98 ± 0.01} & 0.93 ± 0.01 & \textbf{0.01 ± 0.01} & 0.05 ± 0.01 & \textbf{0.08 ± 0.01} & \textbf{0.997409 ± 0.0002} \\
   
  (9) \hspace{0.1cm} $\hat{\mu}_{\text{IT-13B}}$ & \textbf{0.91 ± 0.01} & \textbf{0.98 ± 0.01} & \textbf{0.94 ± 0.01} &0.14 ± 0.01 &0.19 ± 0.01 & 0.16 ± 0.01 & 0.996978 ± 0.0013 \\

  \bottomrule
\end{tabular}
\end{sc}
\caption{Mean and standard deviation classification, calibration, and deferral performance against test split (N=540) on the OSCLMRIC dataset test split.}
\label{clf_performance}
\end{table*}

\begin{figure*}[h]
  \centering
  \includegraphics[width=2.13\columnwidth]{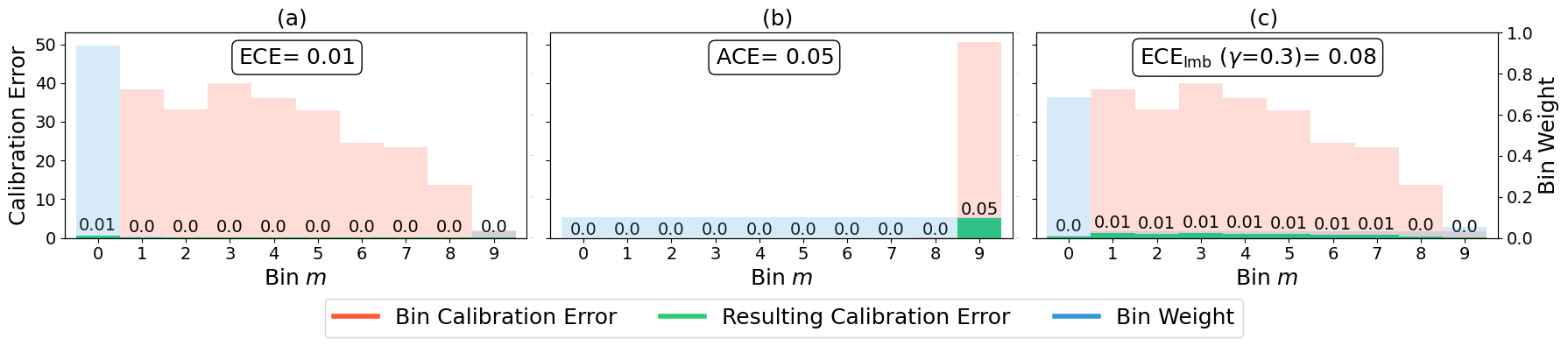}
  \caption{Visual comparison of the effectiveness of the calibration metrics (a) $\mathrm{ECE}$, (b) $\mathrm{ACE}$ and (c) $\mathrm{ECE}_{\mathrm{Imb}}$ of $\hat{\epsilon}_{\text{IT-13B}}$ predictions against the OSCLMRIC dataset test split.}
  \label{ece_imb_lo}
\end{figure*}

\begin{table}[!ht]
\centering
\footnotesize
\begin{sc}
\begin{tabular}{lccc}
  \toprule
  \multirow{2}{*}{\textbf{Setup}} & \multicolumn{3}{c}{\textbf{LLM Efficiency}} \\
  \cmidrule(lr){2-4} &
     \textbf{Rel. s./Gen.}$\downarrow$ & \textbf{Mem.}$\downarrow$ & \textbf{E.R.}$\downarrow$\\
   \midrule

  \hspace{0.1cm} $\text{BASE-13B}$ & \textbf{0.08 ± 0.02} & \textbf{6.02} & 0.29 \\

  \rowcolor{gray!15} \hspace{0.1cm} $\text{BASE-70B}$ & 1.00 & 32.08 & 0.04 \\

  \hspace{0.1cm} $\text{IT-13B}$ & \textbf{0.08 ± 0.02} & \textbf{6.02} & \textbf{0.0002} \\
   
  \bottomrule
\end{tabular}
\end{sc}
\caption{LLM efficiency metrics for selected setups from the OSCLMRIC dataset test split.}
\label{llm_efficiency}
\end{table}

\paragraph{Data.}
We utilise two datasets in our experiments: \textit{OSCLMRIC} and \textit{MIMIC-500}. OSCLMRIC (Oxford Secondary Care Lumbar MRI Cohort), contains 100 professionally radiological reports. The dataset is highly imbalanced, with approximately 95\% of labels negative. Each report is parsed to detect the binary presence of foraminal stenosis (FS), spinal canal stenosis (SCS), and lateral recess stenosis (LRS)—at six lumbar spine levels, resulting in 1,800 examples. We randomly split the data into 30\% for generating an instruction-tuning dataset, 20\% for training the hidden-state classifier, 20\% for validation, and 30\% for testing. Due to the complexity of these reports, we use this dataset as our primary dataset in evaluating the computational efficacy of our method in addition to the pilot study in assessing the effectiveness of our proposed LLM's guidance. Due to privacy reasons, we cannot use this dataset to compare the performance of our methodology against proprietary and state-of-the-art (SOTA) LLMs. Instead, we utilise  Tulu V2 70B as the SOTA baseline model for our experiments against the OSCLMRIC dataset, of which was the highest-performing open-source LLM against several benchmarks at the time these experiments were conducted \cite{open-llm-leaderboard}

For performance comparison against proprietary and SOTA LLMs, we utilise MIMIC-500 \cite{gu2024chexgptharnessinglargelanguage}, a publicly available expert-annotated set of 500 chest X-ray reports (split into impressions and findings sections) of which there exists the SOTA LLM classification benchmarks of GPT-3.5 \cite{brown2020language} and GPT-4 \cite{openai2024gpt4technicalreport}. We focus on the \textit{impressions} section of the reports, as SOTA LLMs performed best against this section. Each report is parsed for 10 classifications, each with high imbalance. To ensure full reproducibility, we disable stochastic processing in the model's decoding stage during experiments with this dataset. MIMIC-500 is a subset of MIMIC-CXR-JPG \cite{mimiccxr}, which includes noisy labels from the predictions of CheXpert \cite{irvin2019chexpert}, an open-source rule-based tool. In instruction-tuning our LLM for MIMIC-500, we greatly restrict our access to data in simulating low-count environments commonly found in healthcare, and utilise just .6\% (1,500) of MIMIC-CXR-JPG (independent of MIMIC-500), with 500 each for instruction-tuning data generation, validation and training the hidden-state classifier. In generating instruction-tuning data, we guard-rail with the noisy labels of CheXpert.

\subsection{Computational Results}
\paragraph{Abbreviations and Evaluation Metrics.}
We abbreviate IT and base (non-IT) models as subscripts ``INSTRUCT" and ``BASE", respectively. Tests for statistical significance (SS) use a one-sided non-parametric Mann-Whitney U test \cite{10.1214/aoms/1177730491}, unless stated otherwise. The Area Under Accuracy-Rejection Curve (AUARC) and F1-score measure deferral and classification performance, respectively. Calibration is evaluated with $\mathrm{ECE}$, $\mathrm{ACE}$, and $\mathrm{ECE_{Imb}}$. LLM efficiency is assessed based on \textit{Rel. s/Gen.} (seconds/generation relative to the baseline), \textit{Mem.} (GPU VRAM in GB), and \textit{E.R.} (error rate: unsuccessful to successful generations).

\paragraph{Imbalanced ECE is an Effective Measure of Calibration.}Figure \ref{ece_imb_lo} shows the limitations of ECE and ACE in imbalanced situations using $\hat{\epsilon}_{\text{IT-13B}}$ predictions. ECE concentrates most bin weight in the first bin, neglecting other bins' calibration errors. ACE improves on ECE but still mainly represents the negative class. Imbalanced ECE resolves this by modifying ECE to reflect calibration errors more globally.

\paragraph{Instruction-Tuned Models Are More Accurate, Equally Calibrated, Better Deferral Systems, and More Efficient.} We find a SS improvement in F1-Score for $\hat{t}_{\text{IT-13B}}$ compared to $\hat{t}_{\text{BASE-70B}}$, demonstrating the success of our instruction-tuning methodology [Table \ref{clf_performance}, Rows (1) \& (4)]. $\hat{\epsilon}_{\text{IT-13B}}$ shows the best deferral performance, surpassing $\hat{\epsilon}_{\text{BASE-70B}}$ [Table \ref{clf_performance}, Rows (8) \& (5)]. $\hat{\mu}_{\text{IT-13B}}$ achieves the highest classification performance [Table \ref{clf_performance}, Row (9)], outperforming $\hat{\epsilon}_{\text{IT-13B}}$ and $\hat{t}_{\text{IT-13B}}$ (\p$<$0.01). This improvement is consistent across all setups, highlighting the distinct value of verbalised and hidden-state sources for classification. There is a positive correlation of $\rho = 0.53 \pm 0.07$ between these predictions. When $\hat{\mu}$ is correct and $\hat{t}$ and $\hat{\epsilon}$ differ, a correct $\hat{\epsilon}$ prediction is associated with an incorrect $\hat{t}$ prediction 75 ± 0.14\% of the time, and vice versa 75 ± 0.25\% of the time. This trend is also seen in MIMIC-500 experiments, where $\hat{\mu}_{\text{IT-13B}}$ achieves the highest macro-averaged F1-score (Table \ref{clf_tab}). Additionally, the IT model significantly improves across all LLM efficiency metrics [Table \ref{llm_efficiency}], producing the fewest errors, requiring the least memory, and taking the shortest time to generate.

\paragraph{The Combined Prediction Has The Greatest Deferral Performance in Mostly-Autonomous Systems.}
We find a SS (\p$<$0.1) improvement in deferring on $\hat{\epsilon}_{\text{IT-13B}}$ and classifying on $\hat{\mu}_{\text{IT-13B}}$ (i.e. taking $\tilde{p} = \hat{\epsilon}$ and $\hat{p} = \hat{\mu}$) compared to deferring and classifying solely on $\hat{\epsilon}_{\text{IT-13B}}$ when deferral is $<$5\%. This suggests that the combined prediction is a better predictor in high-uncertainty situations. For deferral rates $\geq$5\%, there is no significant difference between the two strategies, but both significantly outperform all other setups.

\paragraph{Our Guided Deferral LLM is Competitive With Proprietary SOTA LLMs.}
Table \ref{clf_tab} shows that our guided deferral LLM is competitive with proprietary SOTA LLMs. Despite using only 0.6\% of the training data, it outperforms both in classifying 3 of the conditions. Overall, $\hat{\mu}_{\text{BASE-13B}}$ exceeds the micro and macro-averaged F-score of GPT-3.5 and falls just short of GPT-4, despite significantly smaller numbers of parameters, and with data privacy intact.
\begin{table}[!t]
\centering
\footnotesize 
\begin{sc}
\setlength{\tabcolsep}{1mm}
\newcolumntype{g}{>{\columncolor{gray!15}}c}
\begin{tabular}{lcgcg}

  \toprule 
  \multirow{2}{*}{\textbf{Condition}} & \multicolumn{4}{c}{\textbf{Prediction Source}} \\
  
     \cmidrule(lr){2-5}  & $\hat{t}_{\text{BASE-70B}}$ &
      GPT-3.5 & $\hat{\mu}_{\text{IT-13B}}$ & GPT-4 \\
   \midrule

  Atelectasis & 95.96 &  95.77 & \underline{98.02} & \textbf{98.52}\\

   Consolidation & 85.22 & 85.50 & \underline{98.04} & \textbf{99.01}\\

   Effusion & 87.32  &  90.43 & \underline{96.69} & \textbf{97.69}\\

    Fracture & 76.19 &   \textbf{93.88} & 80.00 & \underline{88.89}\\

    Lung Opacity & 79.88  & 87.16 & 88.17 & \textbf{93.06} \\

   Nodule & 72.34 &  80.95 & \textbf{86.36} & \underline{85.71} \\

  Pleural Lesion & 44.44 &  \textbf{95.08} & 77.78 & \underline{90.00} \\

  Pneumothorax &  81.63  & 60.24 & \textbf{97.78} & \underline{95.45}\\
   
  Pulmonary Edema & 87.13 &  67.24 & \textbf{92.68}  & 85.47\\ 

  *W.M.S & 86.24  & 73.97 & \underline{89.72} & \textbf{94.12}\\

  \hline
    Micro-Avg. & 84.94 & 92.68 & \underline{93.17} & \textbf{93.87}\\
    Macro-Avg. & 79.71 & 89.64 & \underline{90.52} & \textbf{92.79}\\
    \bottomrule 
    Efficient, Effective & \multirow{2}{*}{\xmark}  &  &  \multirow{2}{*}{\cmark} &  \\ AND Open-Source? &  & \cellcolor{gray!15} \raisebox{1.6ex}[1.6ex]{\xmark} & &\cellcolor{gray!15} \raisebox{1.6ex}[1.6ex]{\xmark} \\
    \bottomrule
\end{tabular}
\end{sc}
\caption{\textbf{Our proposed guided deferral LLM is competitive with state-of-the-art proprietary LLMs in classification.} F-scores for classification of chest X-ray conditions on the MIMIC-500 dataset for proprietary LLMs (GPT-3.5 and GPT-4) against our guided deferral IT LLM on the combined probability predictions $\hat{\mu}_{\text{BASE-13B}}$. We additionally include $\hat{t}_{\text{BASE-70B}}$ for inference on the effectiveness of our instruction-tuning methodology. Highest and second-highest performing metrics are in \textbf{boldface} and \underline{underlined}, respectively. SOTA LLM baselines are \colorbox{gray!15}{highlighted in grey}. *W.M.S = Widened Mediastinal Silhouette.}
\label{clf_tab} 
\end{table}

\subsection{Pilot Study Results}
\begin{figure}[t]
\begin{center}
\centerline{\includegraphics[width=0.99\columnwidth]{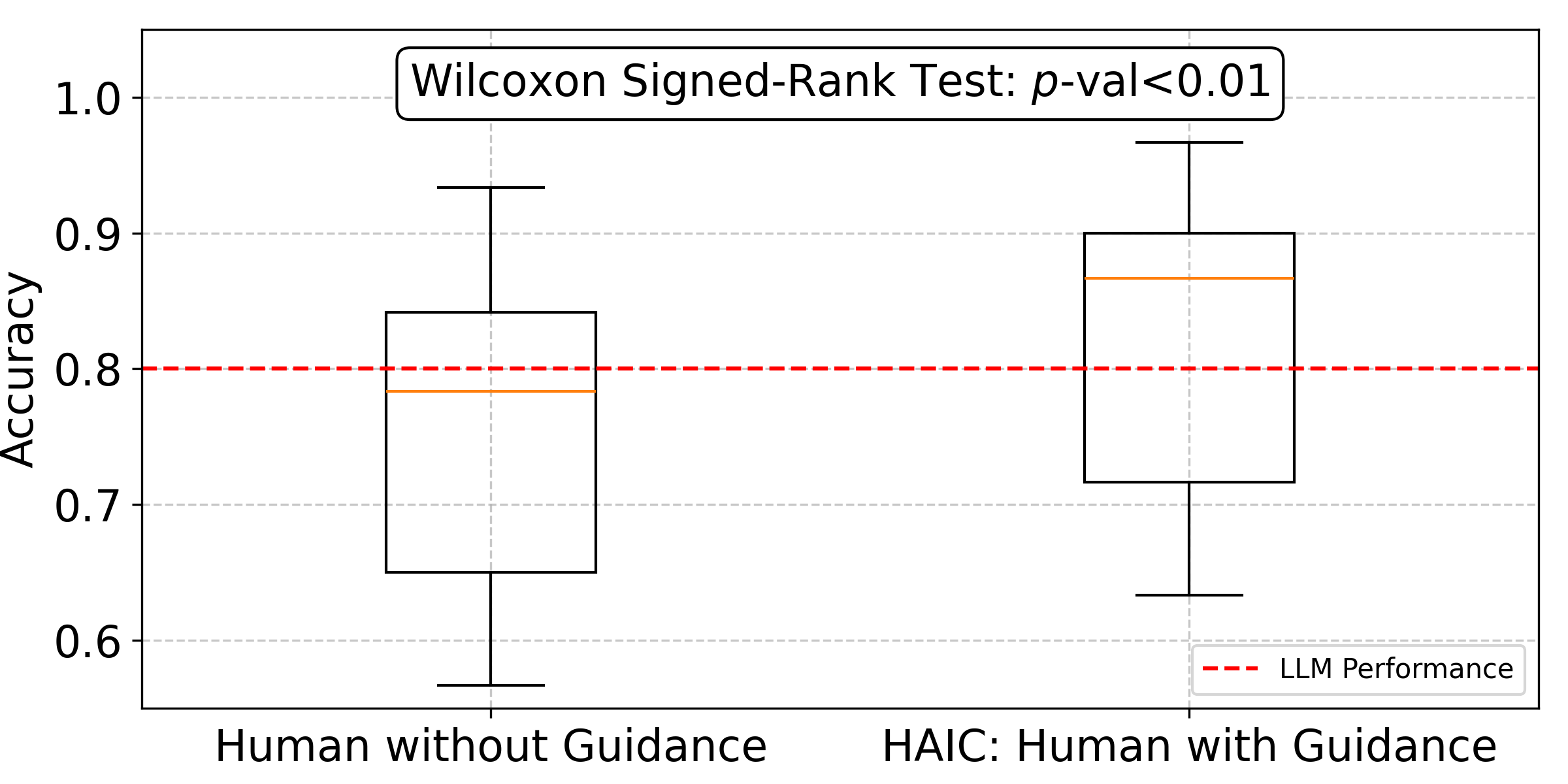}}
\caption{Accuracy of prediction methods from pilot study. On average, guided humans (right box plot) outperform both unguided humans (left box plot) and the LLM (red dashed line) alone.}
\label{boxplot}
\end{center}
\end{figure}

Following the computational deferral results in high uncertainty, we choose $\theta=\tilde{p}_{31}$ with $\tilde{p}=\hat{\epsilon}$ and $\hat{p} = \hat{\mu}$.
\paragraph{Guidance Enhances Human Decision-Making.}
Figure \ref{boxplot} displays the accuracy of participant decision-making with and without guidance, compared to LLM performance.
All participants' accuracy improved with guidance. We rejected the null hypothesis that performance without guidance equals performance with guidance (paired one-sided Wilcoxon signed-rank test, \p$<$0.01). When confronted with AI disagreement, the provided guidance proved effective in assisting participants to arrive at the correct final decision. This was true not only when the participant was incorrect, but importantly also when the LLM was incorrect. We rejected the null hypothesis that the proportion of humans changing their prediction is independent of the AI's prediction correctness ($\chi^2$-test, \p$<$0.01).

\section{Discussion and Conclusion}
Successful AI implementation in healthcare requires combining human and AI strengths, designing systems that are accessible and beneficial to all. Our approach develops a guided deferral framework using LLMs for clinical decisions. We demonstrate that efficient and open-source LLMs can achieve effective deferral and classification performance, competitive with SOTA proprietary systems, with data privacy intact and without the requirement of high-performance compute. We address issues with calibration metrics (ECE and ACE) in imbalanced data commonly found in healthcare tasks, proposing imbalanced ECE as a solution. The effectiveness of our guided deferral system is validated through a pilot study, of which demonstrates complementarity. Our guided deferral system is theoretically adaptable to any data modality or pathology in healthcare, as well as other high-risk domains. However, its practical application requires rigorous ethical evaluation to address the potential risks associated with deploying AI in these contexts. To prevent unexpected errors during implementation, clinicians need thorough education on the system's capabilities. This includes addressing biases such as anchoring and confirmation bias, and informing them about the system's training data distribution, including data drift and out-of-distribution instances.

\section*{Ethical Statement} 
The OSCLMRIC dataset was collected by Professor Jeremy Fairbank with HRA approval (IRAS Project ID: 207858, Protocol Number:12139). The use of AI in decision-critical settings, such as healthcare, raises significant ethical concerns. Our work employs a LLM to provide textual guidance to humans, using instruction-tuning to minimise hallucinations, as validated by our results. However, ethical considerations must address the risk of presenting users with misleading or incorrect information. The environmental impact of LLMs also warrants attention, as these models are computationally intensive and demand substantial energy. While we take steps toward more sustainable applications, current AI practices still carry a notable carbon footprint. Future research should prioritise energy-efficient LLMs and alternative methods to reduce environmental impact. Finally, LLMs pose privacy risks, as sensitive information can potentially be extracted from training data \cite{hu2022membershipinferenceattacksmachine}. In healthcare, where training data may include patient information, strict data sanitisation, privacy-preserving training, and cautious data use are essential to mitigate data leakage.

\section*{Acknowledgements}
JS is supported by the EPSRC Center for Doctoral Training in Health Data Science (EP/S02428X/1). JS gratefully acknowledges the Kellogg College Research Support Grant of which supported this research. The authors acknowledge UKRI grant reference EP/X040186/1 (Turing AI Fellowship: Ultra Sound Multi-Modal Video-based Human-Machine Collaboration). The authors extend their gratitude to Professor. Andrew Zisserman and Dr. Amir Jamaludin from the Visual Geometry Group, University of Oxford, for the provision of this data. A special thank-you to all participants of the pilot study.
\bibliography{aaai25}

\end{document}